\pdfoutput=1

\documentclass[11pt]{article}

\usepackage[final]{acl}

\usepackage{times}
\usepackage{latexsym}
\usepackage{amsmath}
\usepackage{float}

\usepackage[T1]{fontenc}

\usepackage[utf8]{inputenc}

\usepackage{microtype}

\usepackage{graphicx}
\usepackage{makecell}

\title{DNB-AI-Project at SemEval-2025 Task 5: An LLM-Ensemble Approach for Automated Subject Indexing}

\author{Lisa Kluge \\
  Deutsche Nationalbibliothek \\
  Frankfurt am Main, Germany \\
  \texttt{l.kluge@dnb.de} \\  \And
 Maximilian Kähler \\
  Deutsche Nationalbibliothek \\
  Leipzig, Germany\\
  \texttt{m.kaehler@dnb.de} \\
  }

\begin{document}
\maketitle
\begin{abstract}
  This paper presents our system developed for the SemEval-2025 Task 5:
  LLMs4Subjects: LLM-based Automated Subject Tagging for a National Technical
  Library's Open-Access Catalog.
  Our system relies on prompting a selection of LLMs with varying examples of
  intellectually annotated records and asking the LLMs to similarly suggest
  keywords for new records. This few-shot prompting technique is combined with
  a series of post-processing steps that map the generated keywords to the target
  vocabulary, aggregate the resulting subject terms to an ensemble vote
  and, finally, rank them as to their relevance to the record.
  Our system is fourth in the quantitative ranking in the all-subjects track,
  but achieves the best result in the qualitative ranking conducted by subject
  indexing experts.
\end{abstract}

\section{Introduction}

The LLMs4Subject task \cite{dsouza-EtAl:2025:SemEval2025}  
aims at utilising large language models (LLMs) for the task of automated subject indexing on a dataset of open-access publications.
Automated subject indexing is a task that helps enabling access to user-relevant publications by identifying and recording their most important themes and topics in the tagged subject terms.
The ever-growing number especially of digital publications requires reliable
automated systems for this task, which has become infeasible to achieve manually.
In our previous work on automated subject indexing on a similar dataset \cite{kluge-kahler-2024-shot}, we found that the performance of LLMs, while succesfully applied to a range of other tasks \cite{zhao2023survey,yang2024_surveyLLMs,patil2024LLMreview}, was not yet on par with classical supervised machine learning methods. Therefore, it is important to do further research on the capabilities of LLMs in this context.

Rather than fine-tuning models ourselves, the main strategy of our system is to
leverage the existing capabilities of off-the-shelf foundational or
instruction-tuned open-weight LLMs.
In contrast to our previous work, the key contribution of this system is
that it does not rely on only one LLM, but a combination
of different language models along with varying prompts to generate the subject
terms. We found this ensemble approach to dramatically improve the performance
of our system.
To handle the challenge of the controlled vocabulary unknown to the LLMs,
we first generate free keywords with generative LLMs and then map
these onto the vocabulary with a smaller embedding model.

The official quantitative results put us in fourth place, the qualitative results even in first place.
We think that our approach provides valuable insights into the chances and bounds
of the few-shot prompting approach, showing that competitive results are possible
without fine-tuning and large training corpora, simply by combining several
LLMs into an ensemble.

Our code is publicly available.\footnote{\url{https://github.com/deutsche-nationalbibliothek/semeval25_llmensemble}}

\section{Background}

Outlining the field of automated subject indexing, \citet{golub2021automated} presented important fundamentals, approaches and best practices for the task.
Referring to it as index term assignment, \citet{erbs2013bringing} compared and combined two strategies to perform this task: multi-label classification (MLC) and keyword extraction.
Detecting separate strengths, their results aligned with \citet{toepfer2020fusion}, who also found the combination of approaches to be beneficial.

Regarding frameworks for automated subject indexing, the Annif system \cite{suominen2019annif} is an important contribution. Annif has established methods built in, like Omikuji\footnote{\url{https://github.com/tomtung/omikuji}}, which is based on partitioned-label-tree-method Bonsai \cite{khandagale2020bonsai}, or MLLM\footnote{\url{https://github.com/NatLibFi/Annif/wiki/Backend:-MLLM}}, a lexical approach building on \citet{medelyan2009maui}'s Maui.

In earlier work~\cite{kluge-kahler-2024-shot}, we presented experiments with a closed-source LLM on automated subject indexing, but the two baseline methods implemented in Annif mentioned above were found to be as good as or even outperform our LLM-based method.

LLMs have also been utilised for MLC~\cite{peskine2023definitions, d2024context,zhu2023icxml} and keyword extraction~\cite{maragheh2023llm,lee2023toward}.

Recently, building ensembles or fusioning (the results of) LLMs has been addressed as a promising research direction.
There are different works sharing the idea of exploiting the individual strengths and diminishing the weaknesses in different LLMs \cite{jiang-etal-2023-llm,Lu_2023_RoutingToTheExpert,Wang2023_FusingModelsWCompExpertise,Fang_2024_LLM-Ensemble,Wan2024_KnowledgeFusion_FuseLLM}.
Exploring the goal of building ensembles, \citet{Tekin_2024_LLMmTOPLA_Diversity} aimed at maximising diversity and efficiency, whereas \citet{Chen2023_FrugalGPT} targeted the reduction of inference cost.
Not only LLMs have been combined, but also prompts \cite{Pitis2023_BoostedPromptEnsembles,hou2023promptboosting}.
Combining both prompts and models on the task of phishing detection, \citet{Trad_2024_ToEnsembleOrNot_VotingStrategies} contrasted prompt-based ensembles (with one prompt and several LLMs), model-based ensembles and an ensemble consisting of a mixture of prompts and models.

\section{System Overview}

\begin{figure}[t]
  \centering
  \includegraphics[width=.75\linewidth]{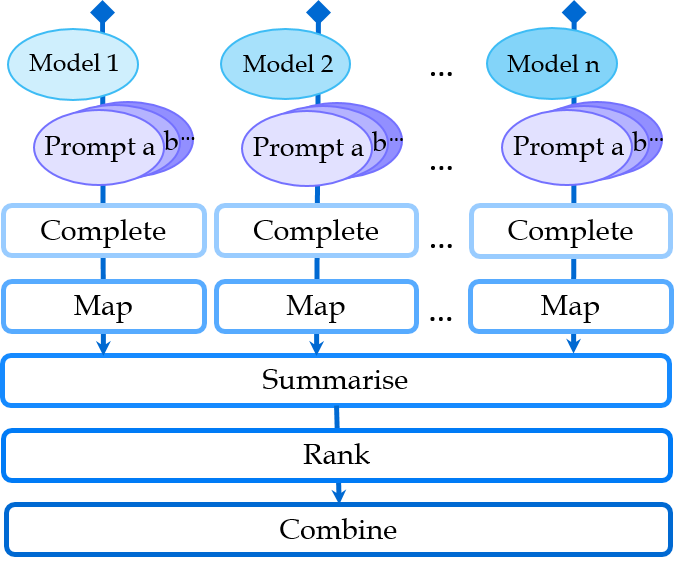}
  \caption{Illustration of our LLM-ensemble approach.}
  \label{fig:method}
\end{figure}

Our system is an enhancement from our previous LLM-based subject indexing approach, described in~\citet{kluge-kahler-2024-shot}. In total, it consists of 5 stages, \emph{complete}, \emph{map}, \emph{summarise}, \emph{rank} and \emph{combine}, as depicted in the overview in Figure~\ref{fig:method}.
At its core, the system approaches the subject indexing task as a keyword generation problem which is solved by a few-shot prompting LLM procedure. As these generated keywords are a priori not restricted to the target vocabulary, a mapping stage with a smaller word embedding model is needed as a supplementary step. In comparison to our previous approach, we have extended the system by combining multiple LLMs and prompts to an ensemble and by introducing an LLM-powered ranking step as in~\citet{d2024context}.

\subsection{Complete}
The first step in our subject indexing system, \emph{complete}, is the generation of keywords following the few-shot paradigm, similar to the procedure in \citet{lee2023toward}.
The \emph{complete}-step is repeated over a range of diverse off-the-shelf open-weight LLMs and prompts with varying few-shot examples.
Our plan and intention of employing a broad variety of models and prompts are twofold. In comparison to a single-model single-prompt setting, we aim to:
\begin{itemize}
  \itemsep-0.4em
  \item Improve recall with an overall greater set of generated subject terms in the ensemble.
  \item Improve precision by utilising the overlap of various model$\times$prompt combinations.
\end{itemize}
Each prompt consists of an instruction
and a set of 8-12 examples illustrating how to perform the subject indexing task with example texts and their gold-standard subject terms.

Details for LLM selection and the composition of the few-shot prompts will be discussed in Sections~\ref{sec:exp_setup_llms} and \ref{sec:exp_setup_prompts}.

\subsection{Map}\label{sec:map}
Keywords generated in the first stage are \emph{mapped} to controlled subject terms
in the target vocabulary using a word embedding model as in~\citet{zhu2023icxml}.

For our \emph{map}-stage, we used \citet{Chen2024_BGEM3}'s BGE-M3-embeddings.
Both generated keywords and target vocabulary are embedded with the same model.\footnote{We embedded the keywords and vocabulary entries without integrating them into a context sentence (which we did in our previous approach).}
To perform nearest neighbour search, we uploaded the embeddings to a Weaviate\footnote{\url{https://weaviate.io/}} vector storage
enabling efficient HNSW-Search~\cite{Malkov2016_HNSW} in $\mathcal{O}(\log L)$ complexity, where $L$ is the vocabulary size.
One feature of the vector storage is that it may be used in a hybrid search
mode~\cite{HybridSearch}, combining vector search and traditional BM25
search~\cite{BM25}.
Thus, each suggested keyword is mapped to the most similar subject term
and the similarity score is stored for later use in the \emph{summarise}-step.
Matches with a low similarity score can be discarded at this stage with a tunable threshold, eliminating
keywords not represented in the vocabulary.

\subsection{Summarise}\label{sec:summarize}
Each prompt and model outputs its own set of predicted subject terms per document after \emph{complete} and \emph{map}.
In the subsequent \emph{summarise}-step, the subject terms are aggregated over all model$\times$prompt combinations by summing the similarities obtained in the \emph{map}-stage (\ref{sec:map}). This score is normalised to a value between 0 and 1 by dividing it by the overall number of model$\times$prompt combinations. Hence, we obtain an ensemble score $s_{\text{ens}}$ for each suggested subject term.
This ensemble score acts as a confidence measure of the individual suggestions and will be included in the final ranking score in the later \emph{combine}-stage (\ref{sec:combine}).

\subsection{Rank}
In the \emph{rank}-stage, that \citet{d2024context} also incorporated in their approach on MLC, another LLM is employed to rank the
subject terms by their relevance. For each predicted
subject term, we ask the model to assess its relevance to the test record at hand on a
scale from 0 (not relevant) to 10 (extremely relevant). Normalised to a value
between 0 and 1, we obtain a relevance score $s_{\text{rel}}$ for each suggested subject term.
Including this additional \emph{rank}-step has two reasons:
Firstly, the relevance score may improve the ensemble score that is, by now, purely based on frequency
and mapping similarity. Asking an LLM to rate the suggestions also
takes into account the context of the text and can thus determine the \textit{relevance}
of the suggestions.
Secondly, this step can be an additional control step for the \emph{map}-stage.

\subsection{Combine}\label{sec:combine}
In the \emph{combine}-stage, a final ranking score for each suggested subject term is obtained as a weighted average from the ensemble and relevance scores.
\begin{equation}
  s_{\text{fin}} = \alpha \times s_{\text{ens}} + (1-\alpha) \times s_{\text{rel}}
  \label{equation:combine}
\end{equation}
In our experiments, we learned setting $\alpha=0.3$ in equation~\ref{equation:combine} resulted in the best ranking (refer to Appendix \ref{appendix:impact_alpha} for more details). In other words, the ordering of the subject terms was best when relying more on the \emph{ranking} than on the \emph{summarisation}.

\section{Experimental Setup}\label{sec:experimental_setup}
\subsection{Data Handling}\label{sec:exp_setup_datahandling}

We used two randomly sampled disjoint subsets ($n=1000$) taken from the union of the development sets given in the all-subjects and the tib-core collection for optimisation and results analysis.
On the first subset, \emph{dev-opt}, we tuned parameters like the model$\times$prompt selection (see Section \ref{sec:exp_setup_ensembleoptimization}) and \emph{combine}-parameter $\alpha$.
The second one, \emph{dev-test}, comprises the data on which we conducted our own evaluation.
In both subsets, we included both English and German texts, as well as all five text types (Article, Book, Conference, Report, Thesis), while keeping the proportions of the overall development set through stratified sampling.

For both input texts and prompts, we used the concatenation of title and abstract as text representation.

\subsection{Vocabulary Adaptation}\label{sec:exp_setup_vocab}

When inspecting early results of our system, we found that the provided vocabulary, GND-Subjects-all, was insufficient to represent the free keywords resulting from the \emph{complete}-stage.
One particular issue was the absence of named entities, that do appear in the full GND but not in this collection.
Plausible keyword candidates, such as country names, are missing and therefore falsely mapped to unrelated subject terms.
Choosing a threshold for minimum similarity between keyword and subject terms was not enough to prevent this kind of error.
Thus, we extended the vocabulary to also include named entities. As the full GND 
would comprise over 1.3 million concepts, we chose to only include
named entities that are actually used in the catalogue of the DNB. In total,
our extended vocabulary includes 309,417 distinct concepts 
(including 200,035 subject terms from the all subjects collection as well as
109,382 named entities from the DNB-catalogue).     
We found that our system produces fewer false positives if we map the named entities cleanly to the extended GND vocabulary and exclude subject terms not belonging to the targeted GND-Subjects-all collection afterwards.
Note that this is also why we only work with the broader GND-Subjects-all vocabulary and
not with the tib-core subset.

\subsection{Language Models}\label{sec:exp_setup_llms}
We experimented with a range of different models for the \emph{complete}-step. We used Llama 3 in 3B-Instruct and 70B-Instruct variants~\cite{dubey2024llama3}, a few versions of Mistral 7B~\cite{jiang2023mistral7b}, Mixtral of Experts~\cite{jiang2024mixtralexperts} and Teuken-7B-Instruct~\cite{ali2024teuken7b}. The overview of models in our final selection is presented in Table~\ref{tab:models}.
We used Llama-3.1-8B-Instruct~\cite{dubey2024llama3} as the ranking model. 

For the \emph{complete}-stage, the number of keywords generated by the LLMs was controlled by setting the minimum number of tokens to 24 and the maximum tokens to 100. 
Find the rest of the hyperparameters affecting the LLMs on our Github\footnote{\url{https://github.com/deutsche-nationalbibliothek/semeval25_llmensemble/blob/main/params.yaml}}.
\begin{table}[t]
  \centering
  \small
  \begin{tabular}{l|l}
    \hline
    \textbf{HF user} & \textbf{Model Name}              \\

    \hline
    \hline
    \scriptsize{\texttt{meta-llama}}       & \scriptsize{\texttt{Llama-3.2-3B-Instruct}}          \\
                     & \scriptsize{\texttt{Llama-3.1-70B-Instruct}}           \\
    \hline
    \scriptsize{\texttt{mistralai}}        & \scriptsize{\texttt{Mistral-7B-v0.1}}                  \\
                     & \scriptsize{\texttt{Mistral-7B-Instruct-v0.3}}         \\
                     & \scriptsize{\texttt{Mixtral-8x7B-Instruct-v0.1}}       \\
    \hline
    \scriptsize{\texttt{teknium}}          & \scriptsize{\texttt{OpenHermes-2.5-Mistral-7B}}        \\
    \hline
    \scriptsize{\texttt{openGPT-X}}        & \scriptsize{\texttt{Teuken-7B-instruct-research-v0.4}} \\
    \hline
  \end{tabular}
  \caption{LLMs used for the completion on the test set.}
  \label{tab:models}
\end{table}

\subsection{Prompts}\label{sec:exp_setup_prompts}
We sampled different sets of prompt examples from the train splits of the all-subjects and tib-core datasets.
To account for the multilinguality of the data, we assembled prompts with only German, only English and mixed-language texts.
However, the gold-standard subject terms that we show to the LLMs are always in German.
Additionally, we also created prompts with a restricted number of subject terms and lemma overlap. Lemma overlap is a measure for similarity between the example text and its subject terms, which we also used in \citet{kluge-kahler-2024-shot}.

Note that we leave the handling of multilinguality completely to the LLMs.
Analysing the keywords resulting from the \emph{complete}-stage on dev-test, it wasn't the case that the models tended to generate English terms, but instead they followed the few-shot demonstrations and output German keywords.

You can view an overview of the prompt example sampling in Appendix \ref{appendix:prompt_instructions}. You can also see the instructions for the \emph{complete}- and \emph{rank}-stages there. The list of examples for each prompt is available on our system's Github\footnote{\url{https://github.com/deutsche-nationalbibliothek/semeval25_llmensemble/tree/main/assets/prompts}}. The templates we used to build the final prompt are also on our system's Github\footnote{\url{https://github.com/deutsche-nationalbibliothek/semeval25_llmensemble/tree/main/assets/templates}}.

\subsection{Ensemble Optimisation}\label{sec:exp_setup_ensembleoptimization}
On the dev-opt subset we ran experiments with 9 models $\times$ 15 prompts,
resulting in 135 sets of subject term suggestions.
However, one cannot expect ever increasing the number of models and prompts to unlimitedly lead to better performance.
Naturally, there is a tipping-point where ensemble performance deteriorates when adding more models or prompts.
Also, there is a trade-off between the number of models and prompts and the computing effort at inference time involed in
the \emph{complete}-step.
Therefore, we conducted an additional optimisation step to find the best subset of models and prompts.

Our optimisation strategy was twofold: In a first Monte-Carlo-like
approach, we repeatedly sampled model$\times$prompt combinations and tested their joint performance as an ensemble,
yielding a subset of 50 out of 135 combinations
that achieve the best precision-recall (PR) balance in terms of area under the precision-recall curve (PR-AUC) on the dev-opt set.
In a second step, we used a chain strategy, where we iteratively removed model$\times$prompt combinations that
did not contribute to the overall performance, narrowing down the selection to 20 combinations.
See Appendix~\ref{sec:ablationstudy_ensembletype} for further results comparing
our ensemble strategy with other strategies as in
\citet{Trad_2024_ToEnsembleOrNot_VotingStrategies}. Also, see the impact of $\alpha$ on the results on the dev-test set in Appendix~\ref{appendix:impact_alpha}.

\subsection{Implementation Details}
We used vLLM \cite{vllm_citation} to serve the LLMs in the \emph{complete}- and \emph{rank}-stages. 
Embeddings for the keywords and the vocabulary were generated using HuggingFace's Text Embeddings Inference\footnote{\url{https://huggingface.co/docs/text-embeddings-inference/index}}. As previously stated, we used Weaviate\footnote{\url{https://weaviate.io/}} as vector storage. 
To create our pipeline and to  manage our experiments, we used DVC \cite{dvc_citation}.

\section{Results}

\subsection{Quantitative Findings}
Table~\ref{tab:quantitative_findings} shows the quantitative results on the all-subjects data of our system and the highest-ranking other teams sorted by averaged recall (R$_{avg}$). In this metric, we are in fourth position.
Note that our approach, in contrast to supervised MLC algorithms,
does not estimate a probability for each subject term in the entire vocab, but rather positively suggests a set of subject terms for each document. Modifying the hyperparameters affecting the number of output tokens can slightly increase the number of different keywords, but our approach doesn't produce result lists of arbitrary length.
For recall@k values with high $k$, the average length of our submitted label lists of 18 makes these scores less adequate to properly estimate our system's performance. Therefore, we also included the scores precision@5 (P$_5$), recall@5 (R$_5$) and F1@5 (F1$_5$) in the table, as we find these metrics to be more insightful to our system's performance.
Figure \ref{fig:pr_curve_quant_ranking} in the Appendix demonstrates how our
system drops off early in recall, while showing competitive results for lower
values of $k$.

\begin{table}[t]
  \centering
  \scriptsize
  \begin{tabular}{l|ccc|c|c}
    \noalign{\hrule height 1pt}
    \textbf{Team} & \textbf{P$_5$} & \textbf{R$_5$} & \textbf{F1$_5$} & \textbf{R$_{50}$} & \textbf{R$_{\text{avg}}$} \\
    \hline
    \hline
    Annif         & \textbf{0.263} & \textbf{0.494} & \textbf{0.343}  & \textbf{0.681}    & \textbf{0.630}            \\
    DUTIR831      & $0.256$        & $0.484$        & $0.335$         & $0.640$           & $0.605$                   \\
    RUC           & $0.230$        & $0.438$        & $0.302$         & $0.642$           & $0.586$                   \\
    icip          & $0.198$        & $0.387$        & $0.262$         & $0.596$           & $0.530$                   \\
    \hline
    Ours          & $0.246$        & $0.471$        & $0.323$         & $0.579$           & $0.563$                   \\
    \noalign{\hrule height 1pt}
  \end{tabular}
  \caption{Official quantitative results for top five teams on the all-subjects task.}
  \label{tab:quantitative_findings}
\end{table}
Looking at the more detailed results for our system (depicted in Appendix \ref{tab:results_language_record_type}), we learned that,
language-wise, one can observe better performance on the German than on the English documents (F1@5=0.332/F1@5=0.307). This could be attributed to the facts that we use a German instruction and that the vocabulary is presented in German.
Potentially, using an English instruction and translating the vocabulary to English - both for the few-shot examples and the mapping stage - would help decrease this gap.
Record-type-wise, Articles are by far the worst category for our system with F1@5=0.157. One reason for this could be the absence of articles in most of our prompts.
All other text types achieve an F1@5 of at least 0.318. Interestingly, Articles are the best record type for the other leading teams, F1@5-wise.

\subsection{Qualitative Findings}

Table~\ref{tab:qual_performance} shows the overall results for the top five teams in the qualitative ranking.
Here, we see our system in the top position. In particular, the evaluation
scenario (case 2) that eliminates those keywords that are technically correct but irrelevant
puts a margin of 2.8\% between our system and the second best team w.r.t. F1@5.
It is unsurprising that the qualitative results are better than the quantitative ones,
as our approach does not involve fine-tuning to the gold-standard.
Subject terms may be helpful and specific in describing the text content,
but at the same time not follow the formal rules applied by TIB's subject specialists when annotating the gold-standard.

\begin{table}[t]
  \centering
  \scriptsize
  \begin{tabular}{l|ccc|c|c}
    \noalign{\hrule height 1pt}
    \textbf{Team} & \textbf{P$_5$} & \textbf{R$_5$} & \textbf{F1$_5$} & \textbf{R$_{20}$} & \textbf{R$_{\text{avg}}$} \\
    \hline
    \hline

    DUTIR831           & 0.488          & 0.316          & 0.384           & 0.611             & 0.485                     \\
    RUC Team           & 0.481          & 0.287          & 0.359           & \textbf{0.618}    & 0.465                     \\
    Annif              & 0.457          & 0.301          & 0.363           & 0.577             & 0.448                     \\
    jim                & 0.404          & 0.287          & 0.335           & 0.545             & 0.426                     \\
    \hline
    Ours               & \textbf{0.526} & \textbf{0.339} & \textbf{0.412}  & 0.615             & \textbf{0.509}            \\
    \noalign{\hrule height 1pt}
  \end{tabular}
  \caption{Official qualitative ranking of the top five teams (\texttt{case 2}).}\label{tab:qual_performance}
\end{table}

Table~\ref{tab:subject_group_f1} in the Appendix shows the F1@5 scores for different subject categories.
Our system was rated particularly high in
architecture, computer science and economics. Worst performance was in history,
traffic engineering and mathematics.

\subsection{Error Analysis}

To get an understanding of the struggles our system faces, we put a small subset
of the dev-test set under manual inspection and compared our system's suggested subject terms
to the gold-standard.
We also analysed the content of title and abstract for these documents.
The questions we had in mind while making this analysis were:
\begin{itemize}
  \itemsep-0.4em
  \item Are there groups of gold-standard subject terms we completely miss?
  \item Are there gold-standard subject terms that are difficult to infer from the given text content?
\end{itemize}
Upon this manual inspection, we noticed that our system benefits from two factors:
specificity of a term and its presence in the concatenated content of title and abstract.
Specific subject terms that are either directly present in the text or are paraphrased in it seem to have the best chance of being correctly predicted.
Generic subject terms are often not found or falsely assigned (e.g. gold-standard: law, found: international law, European law; gold-standard: agricultural policy, found: agriculture). Still, in the list of the most frequent subjects assigned to the dev-test set, there are a lot of general terms, such as history, politics and culture.

Especially when analysing results for the Article text type, which our system performs worst on, we noticed many gold-standard subject terms we suspect to be difficult to directly infer from the given text alone. For example, see the text and its assigned keywords in Appendix \ref{appendix:ablation_study_qualitative_feasibility}. In this record, a lot of words related to the keywords are mentioned in the text (e.g. economic development/growth, agriculture). The exact concepts are not in the text and are also not predicted by our LLM-ensemble. 
Our system relying only on the prompt examples and the concatenation of title and abstract struggles with these types of more complex/abstract relationships between text and subject terms. This is where supervised
learning approaches might have an advantage, as they can learn these relationships from the training data.

Refer to Appendix \ref{appendix:ablation_study_qualitative_feasibility} for more details regarding this error analysis.

\section{Conclusion}

To sum up, we have demonstrated that our ensemble appoach is a promising way to
combine the strengths of different models and prompts, achieving competitive
results in the LLMs4Subjects task.
As we have covered a wide range of prompts and LLMs, we expect our system to
provide a good estimate of the results possible
by prompting LLMs even without fine-tuning.
While our system comes with no extra training costs, a significant drawback is
the high cost involved in prompting multiple LLMs at inference time. 
Appendix \ref{sec:ressources} demonstrates the costs of processing the documents
with each of the LLMs used in our ensemble. Particularly larger models use up
an enourmous amount of GPU-ressources that may be infeasible in productive 
settings.  
In future work, we would like to further investigate techniques for
automated prompt optimisation, such as DSPy \cite{Dspy}, or methods belonging to the family of Parameter-Efficient-Fine-tuning (PEFT). Also we would like to investigate more sophisticated methods for the ensemble combination.

\section*{Acknowledgements} 
We thank the anonymous reviewers for their helpful feedback. 

This work is a result of a research project at the German National Library
(DNB)\footnote{\url{https://www.dnb.de/ki-projekt}}. The
project is funded by the German Minister of State for Culture and the Media
as part of the national AI strategy. With the AI strategy, the German federal
government is supporting the research, development and application of
innovative technologies.

\bibliography{anthology,custom}

\newpage
\appendix

\section{Appendix}
\label{sec:appendix}
\subsection{Prompt Examples and Instructions}\label{appendix:prompt_instructions}
\subsubsection{Prompt Examples Sampling}
\begin{table}[H]
  \centering
  \scriptsize
  \begin{tabular}{lcccc}
    \hline
    & \textbf{Language} & \textbf{N$_\text{examples}$} & \textbf{N$_\text{labels}$} & \textbf{Sim$_{\text{lemma}}$}  \\
    \hline
    1 & German & 8 & random & random \\
    2 & German & 8 & random & random \\
    3 & German & 8 & random & random \\
    4 & German & 8 & random & random \\
    5 & German & 8 & random & random \\
    \hline
    6 & English & 8 & random & random \\
   7 & English & 8 & random & random \\
    8 & English & 12 & random & random \\
    \hline
    9 & Mixed & 8 & random & random \\
    10 & Mixed & 8 & random & random \\
    11 & Mixed & 12 & random & random \\
    \hline
    12 & German & 8 & 1-2 & 0.7-1 \\
    13 & German & 8 & 1-2 & 0-0.3 \\
    14 & German & 8 & 5-10 & 0.7-1 \\
    15 & German & 8 & 5-10 & 0-0.3 \\
    \hline
   
    \hline
  \end{tabular}
  \caption{Prompt sampling overview.}\label{tab:prompts}
\end{table}
\subsubsection{Instruction for \emph{complete}}
The instruction we used for the \emph{complete}-stage:
\begin{quote}
  \small
  Dies ist eine Unterhaltung zwischen einem intelligenten, hilfsbereitem KI-Assistenten und einem Nutzer.
  Der Assistent antwortet mit Schlagwörtern auf den Text des Nutzers.

  \emph{This is a conversation between an intelligent, helpful AI-assistant and a user. The assistant replies with keywords to the text entered by the user.}
\end{quote}

\subsubsection{Instruction for \emph{rank}}
This is the instruction we used for the \emph{rank}-stage:
\begin{quote}
  \small Du erhälst einen Text und ein Schlagwort. Bewerte auf einer Skala von 1 bis 10, wie gut das Schlagwort zu dem Text passt. Nenne keine Begründungen. Gib nur die Zahl zwischen 1 und 10 zurück.

  \emph{You receive a text and a keyword. On a scale from 1 to 10, estimate how well the keyword fits to the text. Do not give reasons. Only reply with the number between 1 and 10.}
\end{quote}

\subsection{Ablation Study Ensemble Strategy}
\label{sec:ablationstudy_ensembletype}

\begin{figure}[!ht]
  \centering
  \includegraphics[width=.95\linewidth]{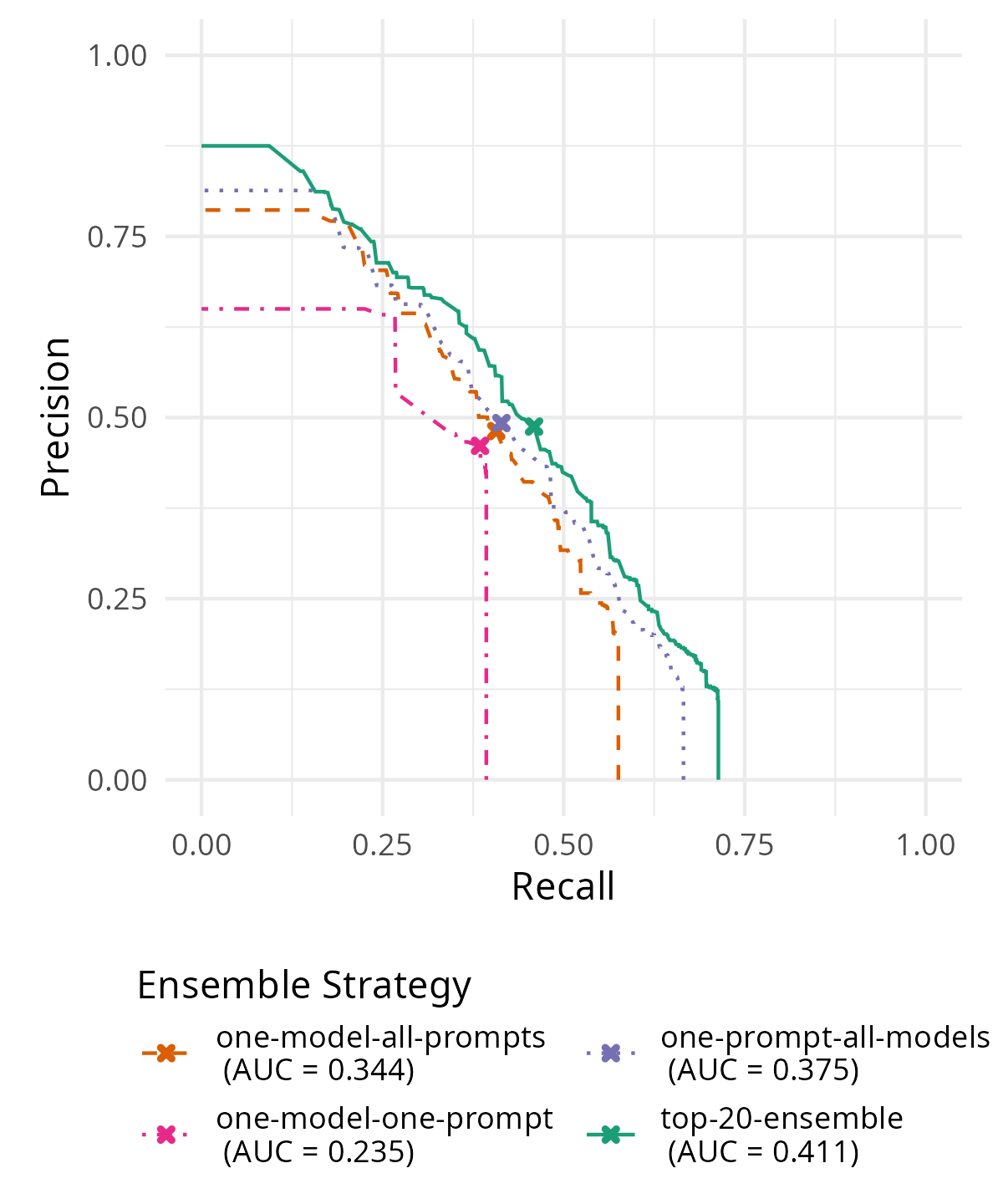}
  \caption{Precision-Recall curves for different model and prompt combinations,
    evaluated as doc-averages over our dev-test set.}
  \label{fig:pr_curves}
\end{figure}

An interesting insight into our system is to evaluate the additional value
of our ensembling approach.
As in \citet{Trad_2024_ToEnsembleOrNot_VotingStrategies}, we complemented the top-20-set
of models$\times$prompt combinations
with other strategies:

\begin{itemize}
  \item \texttt{top-20-ensemble}: with varying models and prompts
        that generate candidates in the complete stage.
  \item \texttt{one-model-all-prompts}: All prompts are used with a single model.
  \item \texttt{one-prompt-all-models}: All models are used with a single prompt.
  \item \texttt{one-prompt-one-model}: A best performing single model-prompt combination is used.
\end{itemize}
All strategies include the rank step and the final combination step as in our
overall system description.
Figure~\ref{fig:pr_curves} shows the PR-curves for the different strategies on the dev-test set. Table~\ref{tab:ensemble_comparison} shows the values
of recall, precision and F1-score that could be obtained with an (F1-)optimal
calibration of the system, also marked with a cross in Figure~\ref{fig:pr_curves}.
Note, unlike the PR-curves in Figure~\ref{fig:pr_curve_quant_ranking},
the curves in Figure~\ref{fig:pr_curves} are not built only on the rank of the
suggested subject terms, but also their confidence scores $s_{\text{fin}}$,
as in Equation \ref{equation:combine}.
Therefore, the curves achieve higher
precision values in comparision to the curves that are built on rank only.

\begin{table}[!ht]
  \centering
  \scriptsize
  \begin{tabular}{l|r|r|r}
    \hline
    \textbf{Ensemble Strategy}                  & \textbf{Precision} & \textbf{Recall} & \textbf{F1} \\
    \hline
    \hline
    \texttt{top-20-ensemble}       & 0.488              & 0.459           & 0.420       \\
    \texttt{one-model-all-prompts} & 0.481              & 0.407           & 0.393       \\
    \texttt{one-prompt-all-models} & 0.492              & 0.414           & 0.407       \\
    \texttt{one-prompt-one-model}  & 0.461              & 0.385           & 0.380       \\
    \hline
  \end{tabular}
  \caption{Precision, recall, F1-score for F1-optimal calibration
    of the system w.r.t. thresholding on confidence scores and
    limiting on rank, computed on dev-test set.}
  \label{tab:ensemble_comparison}
\end{table}

Comparing the precision-recall curves in Figure~\ref{fig:pr_curves}
, we can see that the
\texttt{top-20-ensemble} is well above the other strategies in the high precision as well as the
high recall domain. However, in the part of the curve where the F1-score is optimal, the
ensembling strategies are quite close so that the added value of the ensemble is not as
pronounced compared to the other strategies. This may indicate that the selection
of models and prompts is good (yielding high precision and high recall in the extreme),
but the weighting mechanism of the model-prompt combinations might be improved.
Furthermore, we may conclude that varying the LLMs adds more value to the ensemble
in contrast to varying the prompts.

\subsection{Ablation Study: Single Model Performances}

\begin{figure}[!ht]
  \centering
  \includegraphics[width=.99\linewidth]{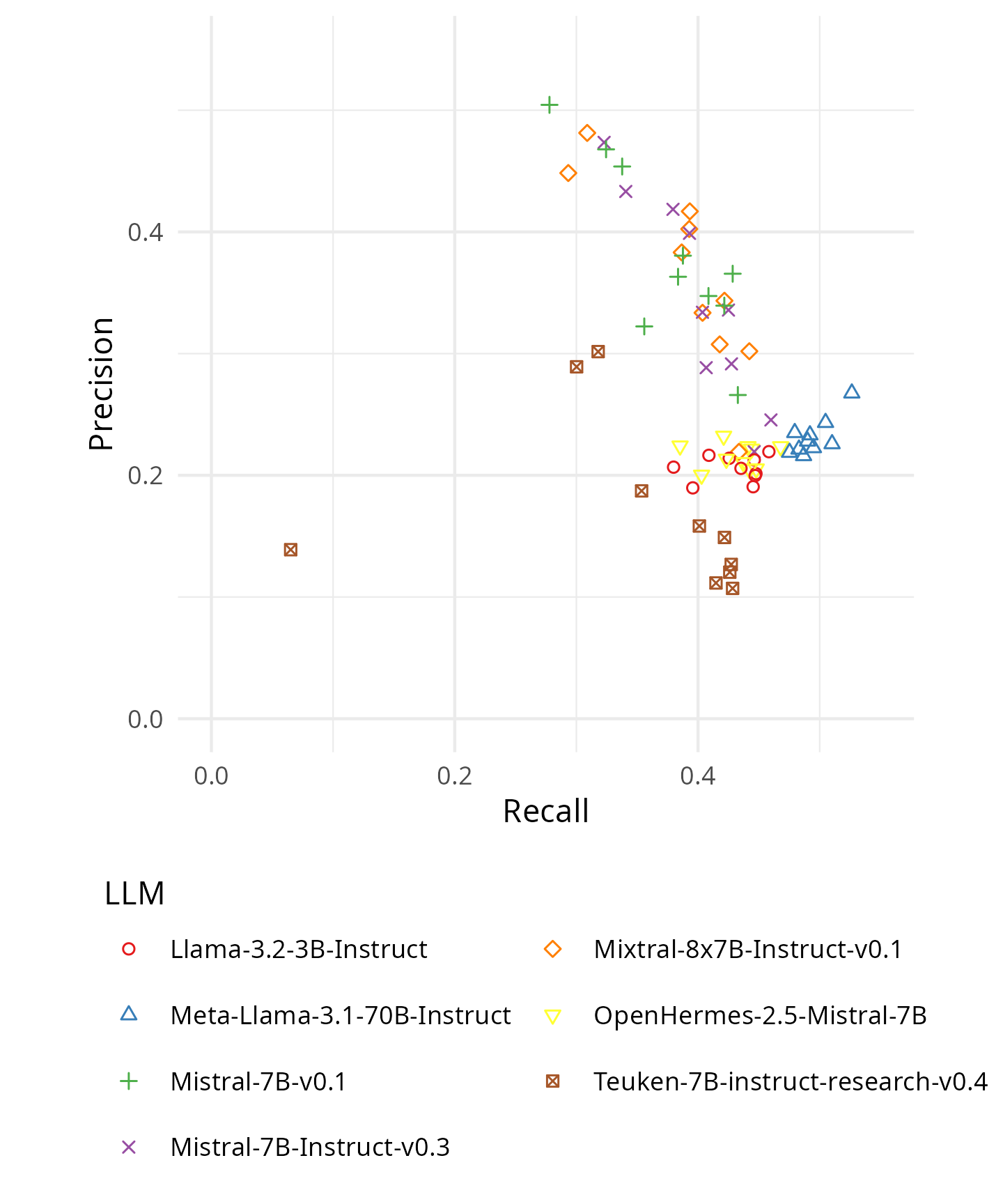}
  \caption{Precision-Recall Balance of single prompt-model combinations on the dev-test sample.}
  \label{fig:single_model_metrics}
\end{figure}

Another interesting insight into our system is how each different
LLM combined with various prompts performs on its own.
To illustrate the spread of precision and recall for the different model$\times$prompt
combinations, see Figure~\ref{fig:single_model_metrics}. These results are
computed on the bare candidate sets suggested by the llm and mapped to
the vocabulary. In this figure, no ranking stage has been applied.

We can see that the most resource-intense model Llama-3.1-70B achieves
highest recall with all prompts. However, precision is not as high as with
the results stemming from the Mistral family.
The Teuken model performs generally worst.
Note, however, that even though a model-prompt combination may
have low performance individually, it may still add value to an
ensemble, as it may provide a different suggestion set than other models.
Indeed, for overall ensemble performance we still found the Teuken model
useful, probably due to its unique tokenizer.

\subsection{Influence of \texorpdfstring{$\alpha$}{α} on PR-AUC}\label{appendix:impact_alpha}
\begin{table}[!ht]
  \centering
  \footnotesize
  \begin{tabular}{l|llll}
    \hline
    \textbf{$\alpha$} & \textbf{1M-1P} & \textbf{1M-AP} & \textbf{1P-AM} & \textbf{top20} \\
    \hline
    0 & \textbf{0.239} & 0.285&0.297&0.301\\
    0.1 & 0.235& 0.344&0.373&0.402\\
    0.2&0.235&\textbf{0.345}&\textbf{0.377}&\textbf{0.411}\\
    0.3& 0.235&0.344&0.375&\textbf{0.411}\\
    0.4& 0.234&0.340&0.369&0.408\\
    0.5&0.232&0.335&0.366&0.405\\
    0.6&0.232&0.333&0.363&0.402\\
    0.7&0.231&0.330&0.359&0.397\\
    0.8&0.230&0.327&0.355&0.394\\
    0.9&0.229&0.324&0.350&0.391\\
    1.0&0.170&0.312&0.328&0.384\\
    \hline
  \end{tabular}
  \caption{PR-AUC scores on the dev-test set for different values of $\alpha$, which determines if the final ranking relies more on the relevance score ($\alpha$<0.5) or the ensemble score ($\alpha$>0.5). The ensembles are abbreviated: one-model-one-prompt (1M-1P), one-prompt-all-models (1M-AP), one-prompt-all-models (1P-AM) and top-20-ensemble (top20).}
  \label{tab:weight}
\end{table}

\subsection{Comparing Precision-Recall Balance among Top Five Teams}

\begin{figure}[H]
  \centering
  \includegraphics[width=.99\linewidth]{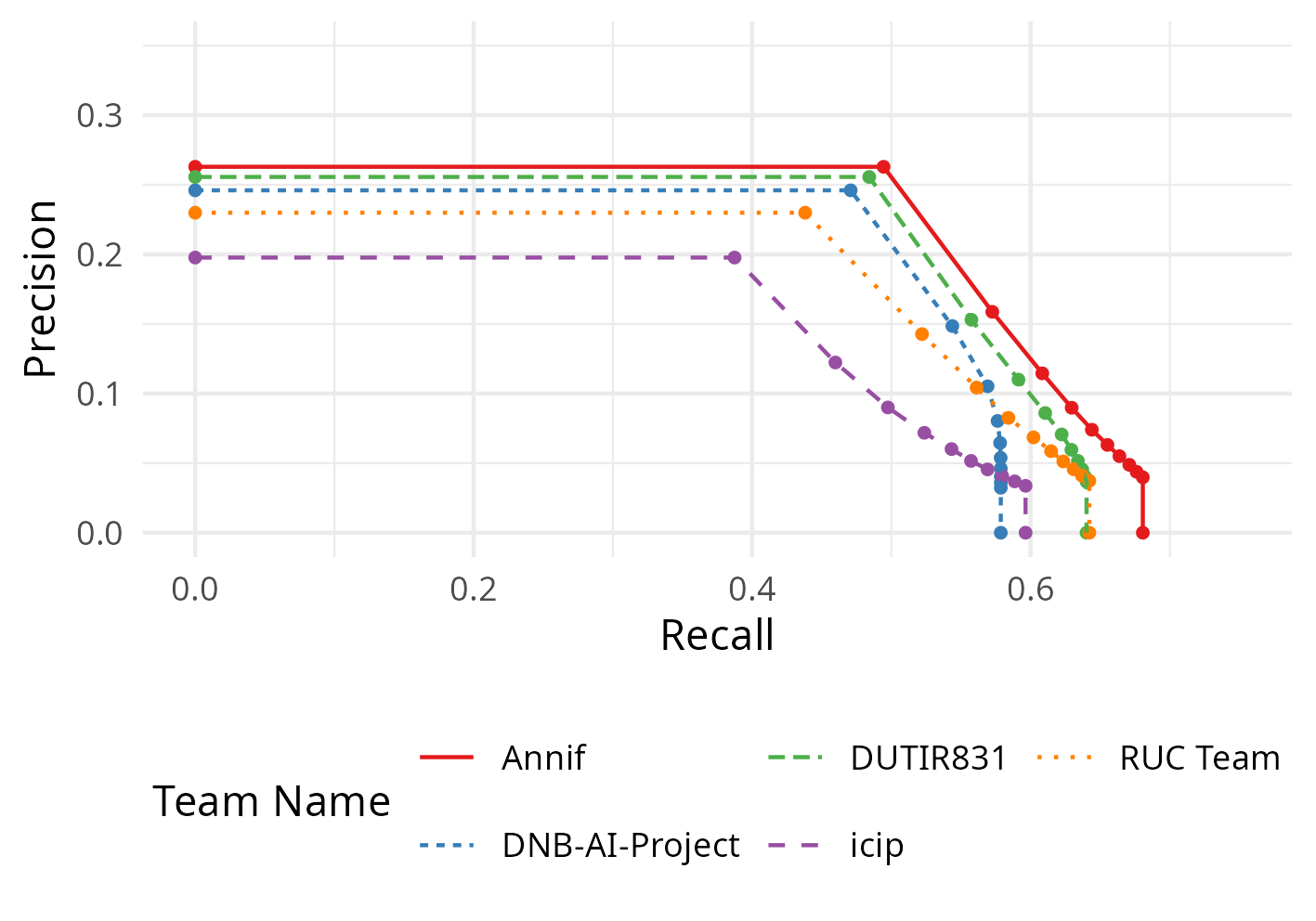}
  \caption{Precision-Recall curves for the top five teams on the all-subjects task.}
  \label{fig:pr_curve_quant_ranking}
\end{figure}

Figure~\ref{fig:pr_curve_quant_ranking} shows the PR curves for
the top five teams on the all-subjects task, plotting the values of precision@k and
recall@k along the increasing values of $k$ as reported in the shared task's
leaderboard.

\newpage

\subsection{Results by Language and Record Type}
\begin{table}[!ht]
  \centering
  \footnotesize
  \begin{tabular}{llll}
    \hline
    \textbf{Record Type} & \textbf{P$_5$} & \textbf{R$_5$} & \textbf{F1$_5$}\\
    \hline
      Article & 0.1108&	0.2685	&0.1569\\
      Book& 0.2396	&0.4898&	0.3218\\
      Conference& 0.2603	&0.4561	&0.3314\\
      Report& 0.2385	&0.4784&	0.3183\\
      Thesis& 0.2912	&0.3932	&0.3346\\
    \hline
    \hline
    \textbf{Language} & \textbf{P$_5$} & \textbf{R$_5$} & \textbf{F1$_5$}\\
    \hline
    de & 0.2545 &	0.4787	&0.3323\\
      en & 0.2307&	0.4566&	0.3065\\
    \hline
  \end{tabular}
  \caption{Metrics precision@5, recall@5 and F1@5 on the test set grouped by record type and language.}
  \label{tab:results_language_record_type}
\end{table}

\subsection{Qualitative Ratings by Subject Category}\label{appendix:qualitative_ratings_by_subject_group}

Table~\ref{tab:subject_group_f1} shows the F1@5-score in the qualitative rating
for each individual subject category. 

\begin{table}[H]
  \centering
  \small
  \begin{tabular}{l|c}
    \hline
    \textbf{Subject Category} & \textbf{F1@5} \\
    \hline
    \hline
    Architecture           & $0.502$       \\
    Chemistry              & $0.428$       \\
    Electrical Engineering & $0.389$       \\
    Material Science       & $0.435$       \\
    History                & $0.322$       \\
    Computer Science       & $0.531$       \\
    Linguistics            & $0.421$       \\
    Literature Studies     & $0.356$       \\
    Mathematics            & $0.343$       \\
    Economics              & $0.486$       \\
    Physics                & $0.357$       \\
    Social Sciences        & $0.409$       \\
    Engineering            & $0.352$       \\
    Traffic Engineering    & $0.343$       \\
    \hline
  \end{tabular}
  \caption{F1@5 scores in the qualitative ranking for different subject categories.}\label{tab:subject_group_f1}
\end{table}

\subsection{Ablation Study: Error Analysis}\label{appendix:ablation_study_qualitative_feasibility}
In addition to the quantitative results, we had a look at $n=50$ random items from the dev-test split. The results are in Table \ref{tab:feasibility}.
\begin{table}[H]
  \centering
  \scriptsize
  \begin{tabular}{c||c|cc||c}
    \hline
                  &                & \multicolumn{2}{c||}{\textbf{Not found}} &                                       \\
    \textbf{Gold} & \textbf{Found} & \textbf{Close}                           & \textbf{Distant} & \textbf{Difficult} \\
    \hline
    140           & 86             & 20                                       & 34               & 44                 \\
                  & (61.4\%)       & (14.3\%)                                 & (24.3\%)         & (31.4\%)           \\
    \hline
    \hline
    26            & 10             & 6                                        & 10               & 17                 \\
                  & (38.5\%)       & (23.1\%)                                 & (38.5\%)         & (64.4\%) \\
  \hline
  \end{tabular}
  \caption{Overview of how many of the gold subject terms in the ablation set are \emph{found}, \emph{not found} but have one or more \emph{close} suggestions, \emph{not found} with only \emph{distant} suggestions found and \emph{difficult}. Bottom row is \texttt{Article}-only.}
  \label{tab:feasibility}
\end{table}

Sample text\footnote{Source: \url{https://github.com/jd-coderepos/llms4subjects/blob/main/shared-task-datasets/TIBKAT/all-subjects/data/dev/Article/en/3A1831638150.jsonld}} to illustrate difficulties of our system with the text type \emph{Article}:
\begin{quote}
  \small
  Chapter 29 Agriculture and economic development "This chapter takes an analytical look at the potential role of agriculture in contributing to economic growth, and develops a framework for understanding and quantifying this contribution. The framework points to the key areas where positive linkages, not necessarily well-mediated by markets, might exist, and it highlights the empirical difficulties in establishing their quantitative magnitude and direction of impact. Evidence on the impact of investments in rural education and of nutrition on economic growth is reviewed. The policy discussion focuses especially on the role of agricultural growth in poverty alleviation and the nature of the market environment that will stimulate that growth. \\
  \textbf{Keywords}: Landwirtschaftliche Betriebslehre (\emph{Agricultural economics}), Agrarpolitik (\emph{Agricultural policy}), Landwirtschaft (\emph{Agriculture}), Wirtschaftstheorie (\emph{Economic theory})
\end{quote}

\subsection{Hardware and Ressources}\label{sec:ressources}

All our computations were run on our internal hardware consisting of
\texttt{2 x Intel(R) Xeon(R) Gold 6338T CPU @ 2.10GHz} processors with
two \texttt{NVIDIA A100} GPUs (each 80GB RAM) attached. Table~\ref{tab:model_performance} 
shows GPU-hours consumed by generating suggestions for the all-subjects test set of 27.987 documents. 

\begin{table}[H]
\centering
\scriptsize
\begin{tabular}{l|r|r|r}
\hline
\textbf{Model Name} & \textbf{Size} & \textbf{GPUh} & \textbf{it/s} \\
\hline
\hline
\texttt{Llama-3.2-3B-Instruct}   & 3B  & 2$\times$ 02:44 h   & 2.84 \\
\texttt{Llama-3.1-70B-Instruct}  & 70B & 2$\times$ 17:36 h  & 0.44 \\
\hline
\texttt{Mistral-7B-v0.1}        & 7B  & 2$\times$ 04:16 h   & 1.82 \\
\texttt{Mistral-7B-Instruct-v0.3} & 7B  & 2$\times$ 03:50 h   & 2.03 \\
\texttt{Mixtral-8x7B-Instruct-v0.1}       & 56B & 2$\times$ 06:28 h  & 1.20 \\
\hline 
\texttt{OpenHermes-2.5-Mistral-7B} & 7B & 2$\times$ 03:36 h  & 2.16 \\
\hline 
\makecell[l]{\texttt{Teuken-7B-instruct-} \\ \texttt{  research-v0.4} } & 7B  & 2$\times$ 02:59 h   & 2.61 \\
\hline
\end{tabular}
\caption{Number of model parameters, GPU hours and iterations per second for different models generating keywords in the \emph{complete} stage. Times measured for generating suggestions for all-subjects test set.}
\label{tab:model_performance}
\end{table}

\end{document}